\documentclass[aoas,preprint]{imsart}
\setattribute{journal}{name}{}
\arxiv{math.PR/1309.7821}
\RequirePackage[OT1]{fontenc}

\usepackage{enumerate,booktabs}
\usepackage{moreverb,amsmath,natbib,multirow}
\usepackage[colorlinks,bookmarksopen,bookmarksnumbered,citecolor=blue,urlcolor=blue]{hyperref}
\setcitestyle{authoryear,open={(},close={)}}

\newcommand\BibTeX{{\rmfamily B\kern-.05em \textsc{i\kern-.025em b}\kern-.08em
T\kern-.1667em\lower.7ex\hbox{E}\kern-.125emX}}

\DeclareMathOperator*{\argmax}{\arg\,\max}

\def\tbM {\tilde{\mathbf{M}}}
\def\bM {\mathbf{M}}

\def\bT {\mathbf{T}}
\def\bx {\mathbf{x}}
\def\bv {\mathbf{v}}
\def\bw {\mathbf{w}}
\def\bX {\mathbf{X}}
\def\bW {\mathbf{W}}
\def\bG {\mathbf{G}}

\def\by {\mathbf{y}}
\def\bz {\mathbf{z}}
\def\tbz {\tilde{\mathbf{z}}}

\newcommand{\bfeps} {\mbox{\boldmath $\epsilon$}}

\newcommand{\bfSigma} {\mbox{\boldmath $\Sigma$}}

\newcommand{\tbfSigma} {\tilde{\mbox{\boldmath $\Sigma$}}}

\newcommand{\tsigma} {\tilde{\mbox{ $\sigma$}}}

\def\ud{{\rm d }}

\def\uN{{\rm N }}

\begin{document}

\begin{frontmatter}
\title{MPBART - Multinomial Probit Bayesian Additive Regression Trees}
\runtitle{Multinomial Probit Bayesian Additive Regression Trees}

\begin{aug}

\author{\fnms{Bereket P.}
	\snm{Kindo}\thanksref{t1,m1}\ead[label=e1]{kindo@email.sc.edu }},
\author{\fnms{Hao}\thanksref{m2} \snm{Wang}\ead[label=e2]{haowang@sc.edu}},
\and
\author{\fnms{Edsel A.} \snm{Pe\~na}\thanksref{t1,m1}
\ead[label=e3]{pena@stat.sc.edu}}

\affiliation{University of South Carolina\thanksmark{m1}, Michigan State University\thanksmark{m2}}

\address{Department of Statistics\\
University of South Carolina \\
Columbia, South Carolina 29208\\
USA\\
\printead{e1}\\
\phantom{E-mail:\  } }

\address{Department of Epidemiology and Biostatistics\\
East Lansing, Michigan 48824
USA\\
\printead{e2}\\
\phantom{E-mail:\  } }

\address{Department of Statistics\\
University of South Carolina \\
Columbia, South Carolina 29208\\
USA\\
\printead{e3}\\
\phantom{E-mail:\  } }

\thankstext{t1}{Research partially supported by NSF Grant DMS1106435 and NIH Grants
R01CA154731, RR17698, and 1P30GM103336-01A1.}

\runauthor{Kindo,Wang, and Pe\~na}

\end{aug}

\begin{abstract}
This article proposes Multinomial Probit Bayesian Additive Regression Trees (MPBART) as a multinomial probit extension of BART - Bayesian Additive Regression Trees \cite{CGM10}. MPBART is flexible to allow inclusion of predictors that  describe the observed units as well as the available choice alternatives. Through two simulation studies and four real data examples, we show that MPBART exhibits very good predictive performance in comparison to other discrete choice and multiclass classification methods. To implement MPBART, we have developed an \texttt{R} package \texttt{mpbart} available freely from \texttt{CRAN} repositories.

\end{abstract}


\end{frontmatter}

\section{Introduction}
\label{section-introduction}

Multinomial probit (MNP) model for discrete choice modeling is often used in economics, market research, political sciences and transportation. It models the choices made by agents given their demographic characteristics and/or the features of the $K > 2$ available choice alternatives. Examples include the study of consumer's purchasing behavior (e.g., \cite{mcculloch2000bayesian, imai2005bayesian}); voting behavior in multi-party elections (e.g., \cite{quinn1999voter}); and choice of different modes of transportation (e.g., \cite{bolduc1999practical}).  Details of the MNP model in which choices depend on predictors in a linear fashion is studied in \cite{mcfadden1973conditional, mcfadden1989method,keane1992note, mcculloch1994exact, nobile1998hybrid, mcculloch2000bayesian,imai2005bayesian, train2009discrete, burgette2012trace} among others.

Among widely used multinomial choice modeling procedures are the multinomial logit model (e.g.,  \cite{mcfadden1973conditional,train2009discrete}) and multinomial probit model (e.g., \cite{mcfadden1989method, mcculloch1994exact, imai2005bayesian}). The former relies on an assumption that a choice outcome is independent of removal (or introduction) of an irrelevant choice alternative while the latter including MPBART does not make this restrictive assumption. 
In the multinomial probit regression framework, it is assumed that each decision maker faced with $K$ alternatives uses a $(K-1)$ vector of latent variables in order to arrive at their choice. Alternative $k$ is chosen if the $k^{\text{th}}$ entry of the latent vector is positive and greater than the other entries, for $k = 1,\ldots, (K-1)$. If none of the entries of the latent vector are positive, then the ``reference'' alternative $K$ is chosen.

MPBART can also be used as a multiclass classification procedure to classify units into one of $K>2$ classes based on their observed characteristics. Multiclass classification is common in many disciplines. In biology, tumors are classified
into tumor sub-types based on their gene expression profiles (e.g., \cite{khan2001classification}). In environmental sciences,  clouds are classified as clear, liquid clouds, or ice clouds based on their radiance profiles (e.g., \cite{LeeWahAck2004}). Other areas of multclass classification applications include text recognition, spectral imaging, chemistry, and forensic science (e.g., \cite{LiEtAl2004,FauCha2006,EveSpi1987, VerEtAl2012}).

The effect of predictors on the response may be linear or non-linear, of much or little significance, and at times magnified with interactions. When such complicated relationships exist, models that use ensemble of trees often provide appealing framework since variable selection and inclusion of interactions are intrinsic in construction of trees. Some popular ``tree-based'' classification methods include CART \cite{Bre1984, quinlan1986induction}, Bayesian CART \cite{CGM98}, random forests \cite{breiman2001random}, and gradient boosting \cite{friedman2001greedy}. There is a gap in the literature for ``tree based'' statistical procedures that directly deal with the MNP model in which choice specific predictors can readily be incorporated. This article, thus, seeks to fill that void using Bayesian tree ensembles for multinomial probit regression.

A newcomer to the ``tree-based'' family is the Bayesian additive regression trees (BART) \cite{CGM10}.  The innovative idea of BART is to approximate an unknown function $f(\bx)$ for predicting a  continuous variable $z$ given values of input $\bx$ using a sum-of-trees model:
$$
f(\mathbf{x}) \approx \sum_{j = 1}^{n_{\text{\tiny T}}} {g \left(\mathbf{x},T_{j},M_{j} \right)},
$$
where $g\left(\mathbf{x},T_{j},M_{j} \right)$ is the $j^{\text{th}}$ tree that consists of sets of partition rules $T_{j}$  and  parameters $M_{j}$ associated with its terminal nodes. Conceptually, the sum-of-trees structure makes BART adaptive to complicated nonlinear and interaction effects, and the use of Bayesian regularization prior on regression trees minimizes the risk of over-fitting.  Empirically, a variety of experiments and applications of BART has confirmed that it has robust and accurate out-of-sample prediction performance \cite{liu2007predictive, CGM10, Abu_Nimeh2008, bonato2011bayesian}.  The standard BART further extends to binary classification problems  and shows competitive classification performance \cite{ZhaEtAl2010,CGM10}.

The success of BART on predicting continuous and binary variables naturally motivates the question of whether the sum-of-trees structure also helps in predicting multinomial choices and classes, thus, we are interested in the utility of the sum-of-trees for discrete choice modeling. We utilize a Bayesian probit model formulation \cite{AC93, mcculloch1994exact,  mcculloch2000bayesian, imai2005bayesian} in conjunction with the idea of sum-of-trees regression to propose multinomial probit Bayesian additive regression trees (MPBART). Through a comprehensive simulation study with various data generating schemes, we find that it is a serious contender in its predictive performance to existing multinomial choice models and multiclass classification methods and that it usually ranks among the topmost  when a nonlinear relationship exists between the predictors and choice alternatives.

A related work to this article is \cite{agarwal2013new}, which utilizes BART for the purpose of satellite image classification. Their multiclass classification procedure combines binary BART and one-versus-all technique of  transforming a multiclass problem to a series of binary classification problems. Our work is different from theirs in that we consider the problem within the traditional multinomial probit regression framework rather than the one-versus-all framework.


The article proceeds as follows. Section \ref{sec:MPBART} formally outlines the multinomial probit model in general and MPBART in particular along with the associated data structure, Section \ref{sec:prior_posterior} delves into the prior specifications and posterior computation for MPBART. Sections \ref{sec:simu} and \ref{sec:realdata} use simulated datasets and real data examples, respectively to illustrate the predictive performance of MPBART. Section \ref{sec:conclusion} closes the article with concluding remarks.

\section{MPBART: Multinomial Probit Bayesian Additive Regression Trees}
\label{sec:MPBART}
Suppose we have a data set $\left(y_i, \bX_i \right) \text{ for } i = 1, \ldots, n$, where
$y_i \in \left\{1,\ldots,K \right\}$ denotes the available choice alternatives and $\bX_i$ the predictors for the $i^{\text{th}}$ observation. We are interested in estimating
the conditional choice probability $p(y_i = j \mid \bX_i)$ for $j = 1, \ldots, K$.   The observed choice $y_i$ can be viewed as arising from a vector of
latent variables $\mathbf{z}_i \in \Re^{K-1}$ \cite{AC93, geweke1994alternative,imai2005bayesian} via
\begin{equation}
y_i(\bz_i) = \left\{ \begin{array}{ll}
j & \textrm{if $\max(\bz_i) = z_{ij}>0$,}\\
K & \textrm{if $\max(\bz_i)<0$,}\\
\end{array} \right. \label{eq:yz}
\end{equation}
for $j = 1, \ldots, (K-1)$, where $\max(\bz_i)$ denotes the largest element of $\bz_i= (z_{i1}, \ldots, z_{i, K-1})'$. The latent vector $\bz_i$ depends on $\bX_i$ as follows:
\begin{equation}
\label{eqn:sum-of-trees}
\bz_i = \mathbf{G}(\mathbf{X}_i; \bT, \bM) + \bfeps_{i} \quad \textrm{ for } i = 1, \ldots, n,
\end{equation}
where $\mathbf{G}(\mathbf{X}_i; \bT, \bM) =  \left( G_1(\mathbf{X}_i; \bT, \bM), \ldots, G_{K-1}(\mathbf{X}_i; \bT, \bM) \right)^{'}$ is a vector of $K-1$ regression functions and $\bfeps_{i} = (\epsilon_{i1},\ldots,\epsilon_{i,K-1})' \sim \textsc{N}(0,\bfSigma).$ 

The predictors for the $i^{\text{th}}$ observation are comprised of two components $\bv_i \text{ and } \bW_i$ (i.e., $\bX_i = \left( \bv_i, \bW_i \right)$ ).
The first component is a vector of $q$ - demographic variables $\bv_i \in \Re^q$ that describe the subject.
The second component $\bW_i = \left(\bw_{i1}, \ldots, \bw_{i(K-1)} \right),$ where $\bw_{ik}\in \Re^{ r},$ is a matrix of $r$ predictors that vary along the choice alternatives in relation to the reference choice. 
For example, in a market research scenario, the price of the choices faced by individuals in a study is a choice specific predictor that varies along alternatives and the difference between the prices of $k^{\text{th}}$ choice and the reference choice $K$ will be part of $\bw_{ik}$, for $k = 1, \ldots, (K-1)$.

The tree splitting rules of the $k^{\text{th}}$  sum of trees
\begin{equation}\label{eq:tree}
G_k(\bX_i; \bT, \bM) = \sum_{j = 1}^{n_{\text{\tiny T}}} {g \left(\bX_i,T_{kj},M_{kj} \right)} \quad \textrm{ for } k = 1, \ldots, (K-1)
\end{equation}
depend on $\bX_i$ through $\bx_{ik} = \left(\bv_i,\bw_{ik} \right)$.
The $j^{\text{th}}$ tree of the $k^{\text{th}}$ sum of trees, $g\left(\cdot,T_{kj},M_{kj} \right)$, consists of $T_{kj}$, a set of partition rules based on the predictor space, and $M_{kj} = \left\{\mu_{kjl}, l = 1, \ldots, b_{kj} \right\}$, a set of parameters  associated with the terminal nodes.
The partition rules $T_{kj}$ are recursive binary splits of the form
$\left\{ x < s \right\} $ versus $\left\{x\geq s \right\}$, where $x$ is one of the predictors that make up $\bx_{ik}$, and $s$ is a value in the range of $x$. The complete set of parameters of MPBART (\ref{eq:yz})--(\ref{eq:tree}) is thus
\begin{equation*}
\left\{\left(T_{kj},M_{kj} \right)_{ k = 1, \ldots, (K-1), j = 1, \ldots, n_T}, \bfSigma  \right\},
\end{equation*}
where $M_{kj}$ denotes the collection of terminal nodes of the $j^{\text{th}}$  tree in the $k^{\text{th}}$ sum-of-trees.

\section{Prior Specification and Posterior Computation}
\label{sec:prior_posterior}
\subsection{Prior Specification}
\label{subsection:prior}
\subsubsection{The $\bfSigma$ prior:}
\label{subsubsection:sigma-prior}

The MNP model specification in \eqref{eqn:sum-of-trees} exhibits a well documented identifiability issue, for example the multiplication of both sides of \eqref{eqn:sum-of-trees} by a positive constant does not alter the implied choice outcome  \cite{keane1992note, mcculloch1994exact, mcculloch2000bayesian, nobile1998hybrid}.
To circumvent this issue, \cite{mcculloch1994exact,mcculloch2000bayesian,imai2005bayesian} among others restrict the first diagonal element of $\bfSigma$ to equal one, while \cite{burgette2012trace} restricts the trace of $\bfSigma$ to equal $K$. We implement the latter.

Consider an augmented latent model 
\begin{equation}
\label{eqn:augm_mdeol}
\tilde{\bz}_i = \mathbf{G}(\mathbf{X}_i; \bT, \tilde{\bM}) + \tilde{\bfeps}_{i},
\end{equation}
where  $\tilde{\bz}_i = \alpha \bz_i$, 
$\tilde{\bfeps}_{i} =  \alpha \bfeps_i$,
, $\tilde{\bfSigma} = \alpha^2 \bfSigma$, $\tilde{M}_{kj} = \left\{ \alpha \mu_{kjl}; l = 1, \ldots, b_{kj} \right\}$ and $\tilde{\bfeps}_{i} \sim \textsc{N}(0,\tilde{\bfSigma})$. Following \cite{imai2005bayesian,burgette2012trace}, we place the prior 
$$p \left(\bfSigma \right) = \int p \left( \bfSigma, \alpha^2 \right) p\left(\alpha^2 \vert \bfSigma \right) d \alpha^2  \propto 
\vert \bfSigma \vert^{-\frac{\left(v + K \right)}{2} } \left(\rm{tr} \left[ S \bfSigma^{-1}\right]\right)^{-\frac{v\left(K -1 \right)}{2}},$$ 
with a restriction $\text{tr}( \bfSigma) = K$; a constrained inverse Wishart distribution induced by $\tbfSigma \sim \textrm{Inv-Wish } \left(\nu, \alpha_0^2 S \right)$ and  $\alpha^2 \vert \bfSigma \sim \alpha_0^2 \textrm{tr} [S \bfSigma^{-1} ] /\chi^2_{v(K)}.$

\subsubsection{The $T_{kj}$ prior:}
\label{subsubsection:tree-prior}
As in \cite{CGM98} and \cite{CGM10}, the prior on a single tree $T_{kj}$ is specified through a ``tree-generating stochastic process'' apriori independent of $\bfSigma$. The tree prior consists of {(i)} the probability of splitting a terminal node, {(ii)} the distribution of the splitting variable if the node has to split, and {(iii)} the distribution of the splitting rule given the splitting variable. For step {(i)}, the probability that a terminal node $\eta$ splits is given by
\begin{equation*}
\frac{\gamma}{\left(1 + d_{\eta} \right)^\beta}, \quad \alpha \in (0,1), \, \beta \in [0, \infty),
\end{equation*}
where $d_{\eta}$ is the depth of the node. 
A small $\gamma$ and a big $\beta$ result in a tree with small number of terminal nodes. In other words, influence of individual trees in the sum can be controlled by carefully choosing $\gamma$ and $\beta$.    For step {(ii)}, the splitting variable is uniformly selected from all possible predictors, representing a prior belief of equal level of importance placed on each predictor. For step {(iii)}, given a splitting predictor, the splitting value $s$ is taken to be a random sample from discrete uniform distribution of the set of observed values of the selected predictor, provided that such a value does not result in an empty partition.

\subsubsection{The $\mu_{kjl}|T_{kj}$ Prior:}
\label{subsubsection:terminal-node-prior}
Given a tree $T_{kj}$ with $b_{kj}$ terminal nodes, the prior distribution on the terminal node parameters is taken to be
\begin{equation*}
\mu_{kjl} \mid T_{kj} {\buildrel iid \over \sim} \uN\left(\mu_k , \tau^2_k \right) \textrm{ for } k=1,\ldots, (K-1).
\end{equation*}
For binary classification problems (i.e., $K=2$),  \cite{CGM10} propose choosing $\mu_1=0$ and $\tau_1 = {3/(r \sqrt{n_T})}$ so that the sum-of-tree effect $\sum_{j = 1}^{n_{\text{\tiny T}}} {g \left(\mathbf{x},T_{1j},\mu_{1j} \right)}$ assigns high probability to the interval $\left(-3,3 \right)$. We extend their method to the multinomial probit setting by assuming $\mu_k=0$ and $\tau_k = {3/(r \sqrt{n_T})}$ for all $k$. The hyper-parameters $r$ and $n_{T}$ play the role of adjusting the level of shrinkage on the contribution of each individual tree. Default values $r = 2$ and $n_{T} = 200$ are recommended by \cite{CGM10} which we also find reasonable in the multinomial probit setup.

\subsection{Posterior computation for MPBART:}
\label{subsubsection:ivd-scheme}

Our posterior sampling scheme relies on the partial marginal data augmentation strategy \cite{van2010marginal}. Marginal data augmentation (MDA) and partial marginal data augmentation \cite{meng1999seeking,imai2005bayesian, van2010marginal,burgette2012trace} introduce a ``working parameter'' that is identifiable given an augmented data, but not identifiable given the observed data. By strategically augmenting the data, MDA and partial MDA result in a computationally tractable posterior distribution and an MCMC chain with improved convergence.

Our posterior computing is accomplished via cycling through the following three steps (for convenience the intermediate draws are flagged with an asterisk).
\begin{itemize}
\item[(i)] Sample from 
$ \left( \bz, \alpha^2 \right) \mid \bT, \bM, \bfSigma,  \mathbf{y} $ by obtaining random draws of 
\newline
$p\{ (\bz_i)_{i=1,\ldots,n} \mid \bT, \bM, \bfSigma,  \mathbf{y} \}$, and $\left(\alpha^*\right)^2 \sim p\left\{\alpha^2 \mid \bfSigma, \bM, \bT, (\bz_i)_{i=1,\ldots,n} \right\} = p\left\{\alpha^2 \mid \bfSigma \right\}$ followed by transforming to obtain $\tbz^*_i = \alpha^* \bz_i$ for all $i$.
\item[(ii)] Sample from 
$\left(\bT, \tbM^* \right) \sim  p\left \{ \bT, \tbM  \mid (\tbz^*_i)_{i=1,\ldots,n}, \bfSigma, \left(\alpha^*\right)^2,  \mathbf{y} \right \}$ followed by recording $\bM = \tbM^* / \alpha^*$.
\item[(iii)] Sample from $ \left(\bfSigma, \alpha^2 \right) \sim p\left \{ \bfSigma, \alpha^2 \mid \bT, \tbM^*, (\tbz^*_i)_{i=1,\ldots,n}, \mathbf{y} \right \}$ by random draws of 

$p\left \{ \tbfSigma^* \mid \bT, \tbM^*, (\tbz^*_i)_{i=1,\ldots,n}, \mathbf{y} \right \}$ followed by transforming $\tbfSigma^* \textrm{ to } (\bfSigma, \alpha^2)$.
\end{itemize}

Our algorithm utilizes a ``partial marginalization'' strategy \cite{van2010marginal} since the working parameter $\alpha^2$ is updated in steps (i) and (iii), but not in (ii) (cf. the marginalization strategy \cite{imai2005bayesian} where the working parameter is updated in every step).

The first part of obtaining a sample from (i) is iterative random draws of truncated normals from the conditional distribution $\bz_{ik} \mid \bz_{i(-k)},\bT, \bM,\bfSigma \sim \uN \left(m_{ik}, \psi_{ik} \right)$ with $\max \left\{0,  \max(\bz_{i(-k)})\right\}$ as a lower truncation point  if $y_i = k$ and as an upper truncation point of if $y_i \neq k$. The conditional first moment and variance ${m}_{ik}, \text{ and } {\psi}_{ik}$ are given by
\begin{equation}
\label{eqn:cond_moments}
\begin{aligned}
{m}_{ik} & =  \bG_k(\bX_i; \bT, \bM) + \sigma_{k(-k)} \bfSigma_{(-k)(-k)}^{-1} \left[\bz_{i(-k)} \ - \ \bG_{(-k)}(\bX_i; \bT, \bM)\right], \text{ and } \\
{\psi}_{ik} & =  \sigma_{kk} - \sigma_{k(-k)} \bfSigma_{(-k)(-k)}^{-1} \sigma_{k(-k)}^{'},
\end{aligned}
\end{equation}
where $\sigma_{k(-k)}$ is the $k^{\text{th}}$ column of $\bfSigma$ that excludes $\sigma_{kk}$ and $\bfSigma_{(-k)(-k)}$ is the matrix $\bfSigma$ that excludes the $k^{\text{th}}$ column and row.

For (ii), we sample $\left(T_{kj},\tilde{M}^*_{kj} \right) \textrm{for } k=1, \ldots, (K-1), j= 1, \ldots, n_{\tiny{T}}$ via the following. Given all the trees and their terminal node parameters but the $j^{\text{th}}$ tree in the $k^{\text{th}}$ sum of trees, $\tbfSigma, \tbz^*_{i(-k)} \text{ and } \left( \alpha^\star \right)^2$, we observe that
\begin{equation}
\label{eqn:psuedo_response_eqn}
\tbz^{\dagger}_{ik}  = g \left(\bX_i,T_{kj},\tilde{M}_{kj} \right) + \tilde{\epsilon}^{\dagger}_{ik}, \quad \tilde{\epsilon}^{\dagger}_{ik} \sim \uN \left(0, \tilde{\psi}_{ik}  \right), \text{ where }
\end{equation}
$\tbz^{\dagger}_{ik}  = \tbz^*_{ik} - \sum_{l \ne j }^{n_{\text{\tiny T}}} {g (\bX_i,T_{kl},\tilde{M}_{kl} )}  -  \tsigma_{k(-k)} \tbfSigma_{(-k)(-k)}^{-1} [\tbz^*_{i(-k)} \ - \ \bG_{(-k)}(\bX_i; \bT, \tbM)]$ and $ \tilde{\psi}_{ik} =  \left( \alpha^*\right)^2\psi_{ik}$. We use the back-fitting algorithm, also used in \cite{CGM10}, to obtain posterior samples of $\left(T_{kj},\tilde{M}^*_{kj} \right)$ by considering \eqref{eqn:psuedo_response_eqn} as the single tree model of \cite{CGM98}. 
Finally, the posterior sample in (iii) is done through a draw from 
$$ \tbfSigma^* \sim \textrm{Inv-Wish} \left(\nu + n, \tilde{S} + \sum_{i=1}^{n}{\left[ \tbz^*_i - \mathbf{G}(\mathbf{X}_i; \bT, \tbM^*) \right]\left[ \tbz^*_i - \mathbf{G}(\mathbf{X}_i; \bT, \tbM^*) \right]^{'} }\right)$$ followed by taking $\alpha^2$ as $\text{tr}( \tbfSigma^*)/K$ and transforming to obtain $\bfSigma = \tbfSigma^* / \alpha^2$.

\subsection{Posterior-based prediction}
\label{subsec:math_formulation}
In our Bayesian setting, predictions of future observations $y^{\star}$ at new values $\bX^{\star}$ are based upon the posterior predictive distribution
$p(y^{\star} \mid \by) = \int p(y^{\star} \mid \bX, \Theta, \by)p(\Theta, \mid \by)\ud \Theta,$ where $\Theta$ consists of all unknown parameters of MPBART.
For a given loss function, predictions of $Y^{\star}$ are made using the optimal choice $a\in \{1,\ldots,K\}$ that minimizes the expected posterior predictive loss:
\begin{align*}
E_{y^{\star} \mid \by}  L\left(y^{\star}, a \right)  & =    \int  L\left(y^{\star}, a \right)   p(y^{\star} \mid \by) \ud y^{\star},
\end{align*}
where $L\left(y^{\star}, a \right)$ is the loss
function of using class $a$ to predict the unknown choice outcome $y^{\star}$.
We assume that the loss function $L(y,a)$ assigns a pre-specified non-negative loss to every combination of action $a \in \{1,\ldots,K\}$ and true choice $y \in \{1,\ldots,K\}$.  These pre-specified loss combinations are described in Table \ref{Table-MisCost} and can equivalently be expressed as
\begin{equation}
\label{eqn-Lossfun}
L\left(y,a\right) =
\sum_{l=1}^{K}{ \sum_{m=1}^{K} { C_{lm} I\left(y=l,a=m \right)}  },
\end{equation}
where $I\left( \cdot \right)$ is the usual indicator function.

\begin{table}[!htb]
\centering
\begin{tabular}{ clllll }
\toprule
&   &\multicolumn{4}{ c }{Prediction $a$} \\ \cline{3-6}
Loss &  & $1$ &$2$ & $\ldots$ & $K$\\
\midrule
{True Choice $y$}  & 1 & $C_{11}$ & $C_{12}$ & $\ldots$ & $C_{1K}$\\
& $2$ & $C_{21}$ & $C_{22}$ & $\ldots$ & $C_{2K}$\\
 & \vdots & \vdots & \vdots & \vdots & \vdots\\
& $K$ & $C_{K1}$ & $C_{K2}$ & $\ldots$ &$C_{KK}$\\ 
\bottomrule
\end{tabular}
\caption{\small Pre-specified costs for the loss function $L(y,a)$.}\label{Table-MisCost}
\end{table}

Under the loss function \eqref{eqn-Lossfun}, the expected posterior predictive loss is then:
\begin{equation}
\label{eqn-Prediction}
E_{y^{\star} \mid \by}  L\left(y^{\star}, a \right) =  \sum_{l=1}^{K}{C_{la} p(y^{\star}=l \mid \by)}.
\end{equation}

We assume that the costs associated with a wrong prediction are all equal to the constant $C$ and correct prediction costs equal to $0$ (i.e., $C_{lm} = C > 0$  for $l \neq m$, and $C_{ll}= 0$). Then the expected posterior predictive loss (\ref{eqn-Prediction}) simplifies to $ E_{y^{\star} \mid \by}  L\left(y^{\star}, a \right)=C \{1-p(y^{\star}=a \mid \by) \},$ which is minimized at
\begin{equation}\label{eq:predictor:u}
a = \argmax_k \{ p(y^{\star}=k \mid \by), k=1,\ldots,K \}.
\end{equation}

The posterior predictive distribution $p(y^{\star}=l \mid \by)$ does not have closed form representation and is thus approximated using Monte Carlo samples drawn from the posterior distributions $p(\Theta \mid \by)$. Once computed, they enable the estimation of the predictions \ref{eq:predictor:u} through a search over the space $a\in \{1,\ldots,K\}$.

\section{Synthetic data examples}\label{sec:simu}
\subsection{A simulation study for multinomial choice model}
\label{sec:sim_mprobit}
In this three choice simulation study, we use a function similar to the one used in \cite{Fri1991} to induce a non-linear relationship between five choice specific predictors $\bw_{k} \in \Re^5, k = 1,2,3$ and the choice alternatives. The choice specific predictors are from i.i.d $\textsc{Unif}[0,1]$.
In addition, we include a predictor $v {\buildrel iid \over \sim} \textsc{Unif}[0,2]$ that describes the observed unit.
Suppose that $ f \left( \mathbf{u} \right) =  20 \sin ( \pi  u_{1} u_{2} ) - 20 ( u_{3}- 0.5 )^2 + 10 u_{4} + 5 u_{5},$  $g\left( v \right) = 8 v$, and
\begin{equation}
\label{eqn:simu_friedman}
\begin{bmatrix}
z_1\\
z_2
\end{bmatrix}
=
\begin{bmatrix}
f(\bw_1 - \bw_3) + g\left( v \right) \\
f(\bw_2 - \bw_3) + g\left( v \right)
\end{bmatrix}
+ \bfeps,  \quad 
\bfeps \sim 
\uN \left( \mathbf{0}, \begin{bmatrix}
1 & 0.5\\
0.5 & 1
\end{bmatrix} \right).
\end{equation}
The response variable is then recorded using
\begin{equation*}
y(\bz) = \left\{ \begin{array}{ll}
k & \textrm{if $\max(\bz) = z_{k}>0$,}\\
3 & \textrm{if $\max(\bz)<0$,}\\
\end{array} \right. \textrm{for } k = 1,2.
\end{equation*}

This true model contains linear, nonlinear, and interaction effects, making it interesting benchmarking dataset. We are mainly interested in how well MPBART is able to predict the choices on a test data. Hence, we simulate a training and test data sets of $500$ observations each and compare the predictive performance on the test data for MPBART, Bayesian multinomial probit model (Bayes-MNP) \cite{imai2005bayesian}, the Multinomial logit (MNL) model \cite{train2009discrete,mcfadden1973conditional}, and the following multiclass classification procedures:  support vector machines  with linear (SVM-L) and radial (SVM-R) kernels \cite{cortes1995support, vapnik1999nature},
random forest (RF) \cite{breiman2001random},
linear discriminant analysis (LDA) and
quadratic discriminant analysis (QDA) \cite{duda2012pattern, friedman2001elements},
multinomial logistic regression (MNL) \cite{mcfadden1973conditional}, classification and regression trees (CART) \cite{Bre1984, quinlan1986induction},
neural networks (NNET),
$K$-nearest neighbors (KNN) \cite{fix1952discriminatory, cover1967nearest} and One vs. All BART (OvA-BART) \cite{agarwal2013new}. we note that for the multiclass classification procedures, a choice specific predictor makes up three separate predictors, one describing each of the choices, putting the total number of predictors for this simulation study at sixteen. For each competing procedure and MPBART,
we selected the tuning parameters via a $10$-fold cross-validation based on the training data.  Table  \ref{tab:listclassifiers} lists
the names of  these competing procedures, the corresponding R-packages utilized and tuning parameters.

\begin{table}[!htb]
\centering
\begin{tabular}{lll}
\toprule
Procedure	&	R Package	&	Tuning parameter(s)	\\
\midrule
RF	&	randomForest	&	mtry	\\
CART	&	rpart	&	no tuning parameters	\\
SVM-L	&	kernlab	&	$C$	\\
SVM-R	&	kernlab	&	$C$ and  $\sigma$	\\
QDA	&	MASS	&	no tuning parameters	\\
LDA	&	MASS	&	no tuning parameters	\\
NNET	&	nnet	&	size and decay	\\
MNL	&	mlogit	&	no tuning parameters	\\
KNN	&	caret	&	$k$	\\
OvA-BART & dbarts & k, power, base\\
\bottomrule

\end{tabular}

 \caption{\small
List of competing classifiers, the R packages utilized, and tuning
 parameters that are chosen by cross-validation.  The abbreviations in the first column stand for
 the procedures mentioned in the second paragraph of Section \ref{sec:sim_multiclass}.}\label{tab:listclassifiers}.
\end{table}

The comparison metric we use in this example and all that follow is test error rate  
\begin{equation}
\label{eqn:test_err_rate}
\frac{1}{m}\sum_{i=1}^{m}{I \left( \hat{y}_i \neq y_i \right)},
\end{equation}
where $y_i$ and $\hat{y}_i$  are the actual and predicted classes for the $i^{\text{th}}$ observation in a given test data set of size $m$. This metric makes use of the loss function in \eqref{eqn-Lossfun} with a misclassification cost of $C_{lm} = 1$ and a cost of $C_{ll} = 0$ for a correct prediction. As can be seen from Table  \ref{tab:friedman_results}, MPBART exhibits a very good out-of-sample predictive accuracy. This is not surprising given the data generating scheme with nonlinear effects.

\begin{table}[!htb]
\centering
\begin{tabular}{|l|c|c|c|c|}
\hline
 \multirow{2}{*}{Procedure} & \multicolumn{2}{c|}{Simulation Study - I} & \multicolumn{2}{c|}{Waveform Recognition}\\ \cline{2-5}
 & Test Error Rate & Rank & Test Error Rate & Rank\\
\hline
MPBART & 0.2725 (0.0060) & 1 & 0.1589 (0.0047)  & 2	\\
Bayes-MNP & 0.3976 (0.0065)  & 7 & 0.2167 (0.0197) & 11	\\
MNL & 0.3921 (0.0064)  & 6 & 0.1721 (0.0052)  & 5	\\
RF  & 0.4023 (0.0059)  & 8 & 0.1676 (0.0043)  & 3	\\
CART  &  0.4791 (0.0080)  & 12 & 0.3113 (0.0068)  & 12	\\
SVM-L & 0.4072 (0.0058)  & 9 & 0.1844 (0.0043)  & 6	\\
SVM-R & 0.3254 (0.0057)  & 3 & 0.1708 (0.0053)  & 4	\\
LDA & 0.4095 (0.0064)  & 10 & 0.1997 (0.0048)  & 8	\\
QDA & 0.3381 (0.0045)  & 4 & 0.2125 (0.0043)  & 10	\\
NNET  & 0.2917 (0.0065)  & 2 & 0.2012 (0.0071)  & 9	\\
KNN & 0.4195 (0.0070)  & 11 & 0.1847 (0.0048)  & 7	\\
OvA-BART  & 0.3908 (0.0059)  & 5 & 0.1550 (0.0035)  & 1	\\
\hline
\end{tabular}
 \caption{\small
Comparison of MPBART, and the procedures listed in Table \ref{tab:listclassifiers} on the first simulation study generated via \eqref{eqn:simu_friedman} and the waveform recognition example \eqref{eqn:waveform}. Training and test data sets of each $500$ observations are used for the first simulation study. Training and test data sets of $300$ and $500$ observations, respectively are used for the waveform recognition example. Average test error rates (with standard errors in parentheses) are reported on 20 replications. }
\label{tab:friedman_results}
\end{table}

\subsection{A simulation study for multiclass classification}
\label{sec:sim_multiclass}
In this simulation study the waveform recognition problem in \cite{Bre1984}, often used as a benchmark artificial data in multiclass classification studies (e.g., \cite{Gam2003, HasTib1996, Kee2005}), is employed.
The model has 21 predictors and a multiclass response with 3 classes. For each observation, the $i^{\text{th}}$ predictor $x_i$ is generated from
\begin{equation}
\label{eqn:waveform}
x_i = \left\{ \begin{array}{ll}
u \, h_1(i) + (1-u )h_2(i) + \epsilon_i, & \textrm{if $y=1$,}\\
u \, h_1(i) + (1-u )h_3(i) + \epsilon_i, & \textrm{if $y=2$,}\\
u \, h_2(i) + (1-u )h_3(i) + \epsilon_i, & \textrm{if $y=3$,}\\
\end{array} \right.
\end{equation}
where $i=1,\ldots,21$, $u \sim \textsc{Unif}[0,1]$, $\epsilon_i \sim \uN\left( 0,1 \right)$, and $h_i$ are three waveform functions: $h_1(i) = \max(6 - |i - 11|, 0)$, $h_2(i) = h_1(i - 4)$, and $h_3(i) = h_1 (i + 4)$.

We generate 20 replications of training and testing data sets with
$300$ and $500$ observations, respectively from \eqref{eqn:waveform} and compare MPBART with classifiers listed in Table \ref{tab:listclassifiers}. Our choice of sample sizes is the same as those in \cite{HasTib1996} so the results can be compared with them.  Table \ref{tab:friedman_results} summarizes the average error rates and standard errors in parentheses based on 20 simulations.  For LDA, QDA and CART, the error rates are consistent with those reported in Table 1 of \cite{HasTib1996}.  MPBART is among best for this data generating scheme exhibiting low test error rates. Note that \cite{HasTib1996} report an error rate of $0.157$ on test data sets achieved by penalized mixture discriminant analysis.

\section{Real data examples}
\label{sec:realdata}
\subsection{Multinomial Choice Example Datasets}
Two discrete choice datasets, fishing mode and travel mode choice datasets, are used to illustrate MPBART. Fishing mode choice data is a survey of 1,182 individuals  who reported their most recent saltwater fishing modes as either ``beach'', ``pier'', ``boat'' or ``charter''. The choice specific variables in this data set are expected catch rates per hour and price of each mode of fishing, while the individual specific predictor is individual's monthly income. Details of this data are in \cite{kling1996implications,herriges1999nonlinear} and we use the version of data available in the R package \textit{mlogit}. The second data records the choice of travel mode between Sydney and Melbourne, Australia as either ``air'', ``train'', ``bus'' or ``car'' \cite{greene2003econometric,kleiber2008applied}. It includes 210 individuals'  choice of travel and the following choice specific predictors: general cost associated with the travel mode choice, waiting time at a terminal (with zero recorded for a travel choice of ``car''), cost of travel mode and travel time. In addition, the individual specific predictors logarithms of household income, and traveling party size are used. We use the version of the dataset in the R package {\it AER} \cite{kleiber2008applied}.

After splitting the fishing mode data into ten and the travel mode data into five nearly equal random folds, we implement the procedures MPBART, Bayesian multinomial probit model (Bayes-MNP), the Multinomial logit (MNL)  and the multiclass classification procedures listed in Table \ref{tab:listclassifiers} with one fold of the data set aside as a test data and the remaining folds utilized for training the models. Table \ref{tab:fishing_mode_results} reports the average test error rates along with their standard errors. MPBART is again among the procedures with the lowest error rates.
\begin{table}[!htb]
\centering
\begin{tabular}{|l|c|c|c|c|}
\hline
 \multirow{2}{*}{Procedure} & \multicolumn{2}{c|}{Fishing Mode} & \multicolumn{2}{c|}{Travel Mode}\\ \cline{2-5}
 & Test Error Rate & Rank & Test Error Rate & Rank\\
\hline
MPBART	&	0.3960 (0.0160)	& 1 & 0.0571 (0.0086) & 2 \\
Bayes-MNP	&	0.5546 (0.0171) & 10 &	0.3286 (0.0394) & 10\\
MNL	&	0.5600 (0.0160) & 11 &	0.3143 (0.0332) &9\\
RF &	0.4746 (0.0148) & 3&	0.0429 (0.0089) &1\\
CART&	0.5372 (0.0147) & 8&	0.1048 (0.0161) &3\\
SVML&	0.5034 (0.0139) & 6&	0.2143 (0.0345) &7\\
SVMR&	0.4882 (0.0194) & 4&	0.1381 (0.0254) &5\\
LDA&	0.4975 (0.0193) & 5&	NA &\\
NNET&	0.5211 (0.0064) & 7&	0.3048 (0.0739) & 8\\
KNN&	0.5406 (0.0189) & 9&	0.1810 (0.0358) &6\\
OvA-BART&	0.4434 (0.0144)  & 2&	0.1143 (0.0158) &4\\
\hline
\end{tabular}
\caption{\small
Comparison results on the fishing mode and choice of travel mode datasets. Classification error rates (with standard errors in parentheses) are reported.}\label{tab:fishing_mode_results}.
\end{table}

\subsection{Multiclass Classification Example Datasets}
Forensic glass and vertebral column classification datasets, both of which are publicly available at the University of California at Irvine (UCI) machine learning data repository \cite{Bache+Lichman:2013}, are used to illustrate MPBART as a multiclass classification procedure in comparison to the multiclass classification procedures listed in Table \ref{tab:listclassifiers}.  The forensic glass classification data set consists of 9 features collected on 214 glass samples, each of which is
classified as one of the 6 glass types: building windows float processed, building windows non-float
processed, vehicle windows float processed, containers, tableware, or
headlamps. The vertebral column data contains 310 patients diagnosed either as normal, having Disk Hernia or Spondylolisthesis. The major function of the human vertebral column is the protection of the spine. It also serves as the body's support system and enables movement by transferring weight muscles connected to it. This dataset records the pathology of the vertebral column and its dependence on the characteristics of the pelvis and lumbar spine. Further detail on the dataset is available in \cite{da2011diagnostic,calle2013computer}. 

In our analysis, we split the forensic glass and vertebral column  datasets into five and ten nearly equal random folds, respectively. One fold of the datasets is set aside as test data and the classification methods in Table \ref{tab:listclassifiers} and MPBART are trained on the remaining folds. Table \ref{tab:forensic_results} shows the average classification error rate with standard errors in parenthesis. QDA could not be implemented in this data set since the representation of observations classified as tableware is very small. For the same reason, we only considered five-fold partitioning of the forensic glass data. MPBART, RF and OvA-BART are the top performing procedures in terms of having the lowest classification error.

\begin{table}[!htb]
\centering
\begin{tabular}{|l|c|c|c|c|}
\hline
 \multirow{2}{*}{Procedure} & \multicolumn{2}{c|}{Vertebral Column } & \multicolumn{2}{c|}{Forensic Glass}\\ \cline{2-5}
 & Test Error Rate & Rank & Test Error Rate & Rank\\
\hline
MPBART	&	0.1466 (0.0324) 	&1	&	0.2946 (0.0182) & 2 \\
RF	&	0.1645 (0.0265) 		&4	&	0.2056	(0.0089) & 1 \\
CART	&	0.1839 (0.0160)		&8	&	0.3272	(0.0356) & 5\\
SVML	&	0.1484 (0.0285)		&2	&	0.3741	(0.0294) & 8\\
SVMR	&	0.1742 (0.0216)		&6	&	0.3086	(0.0222) & 4 \\
LDA	&	0.1968 (0.0335)			&0	&	0.3833	(0.0145) & 9\\
QDA	&	0.1548 (0.0254)			&3	&	NA & NA\\
NNET	&	0.2161 (0.0259)		&10	&	0.3740 (0.0172) & 7\\
MNL	&	0.6129 (0.0304)			&11	&	0.3834	(0.0269)& 10\\
KNN	&	0.1806 (0.0334)			&7	&	0.3506	(0.0316)& 6\\
OvA-BART	&	0.1645 (0.0282)	&5	&	0.3083	(0.0196)& 3\\
\hline
\end{tabular}
 \caption{Classification error rates and 
 standard errors (in parentheses) for vertebral column and forensic glass  data sets.  }
\label{tab:forensic_results}
\end{table}

\section{Conclusion}
\label{sec:conclusion}
We have proposed and tested through simulations studies and real data examples the utility of Bayesian ensemble of trees for Multinomial Probit regression and multiclass classification. Regression trees and their ensembles are widely used for the purpose of classification. However, their use in multinomial probit regression which allows the introduction of choice specific predictors is less explored. MPBART fills that gap in the literature. It exhibits very good predictive performance in a range of examples and is among the best when the relationship between the predictors and choice response is nonlinear. The software implementation of MPBART is freely available as an \texttt{R} package \texttt{mpbart}. For the simulation studies and real data examples, the MPBART tuning parameters selected via cross-validation are available at \url{https://github.com/bpkindo/mpbart_cv_selection/}.

\section*{Acknowledgments}

The authors thank Professor Edward I. George for his 2012 Palmetto Lecture at the University of South Carolina, which partly motivated this research. The
authors also thank Professor James Lynch and Professor Edsel Pe\~na's research group (A.K.M Rahman, Lillian Wanda, Piaomu Liu) for their comments and discussions.

\bibliography{BibMulticlass}

\end{document}